\def\paperID{11751}
\def\confName{ICCV}
\def\confYear{2025}

\def\paperTitle{You Share Beliefs, I Adapt: Progressive Heterogeneous Collaborative Perception}

\def\authorBlock{
    Hao Si \qquad
    Ehsan Javanmardi \qquad
    Manabu Tsukada \thanks{Corresponding author} \\
    The University of Tokyo \\
    {\tt\small \{si-hao, ejavanmardi, mtsukada\}@g.ecc.u-tokyo.ac.jp}
}

\newif\ifreview 
\newif\ifarxiv \newcommand{\arxiv}{\arxivtrue}
\newif\ifcamera 
\newif\ifrebuttal 

\arxiv 

\pdfoutput=1
\documentclass[10pt,twocolumn,letterpaper]{article}
\ifreview \usepackage[review]{cvpr} \fi
\ifarxiv \usepackage[pagenumbers]{cvpr} \fi
\ifrebuttal \usepackage[rebuttal]{cvpr} \fi
\ifcamera \usepackage{cvpr} \fi

\usepackage{graphicx}	
\usepackage{amsmath}	
\usepackage{amssymb}	
\usepackage{booktabs}
\usepackage{times}
\usepackage{microtype}
\usepackage{epsfig}
\usepackage{caption}
\usepackage{float}
\usepackage{placeins}
\usepackage{color, colortbl}
\usepackage{stfloats}
\usepackage{enumitem}
\usepackage{tabularx}
\usepackage{xstring}
\usepackage{multirow}
\usepackage{xspace}
\usepackage{url}
\usepackage{subcaption}
\usepackage{xcolor}
\usepackage[hang,flushmargin]{footmisc}

\usepackage{algorithm}
\usepackage{algpseudocode}

\ifcamera \usepackage[accsupp]{axessibility} \fi

% ...

\ifarxiv  \fi

\newcommand{\R}[1]{{%
    \textbf{%
        \ifstrequal{#1}{1}{\textcolor{red}{R#1}}{%
        \ifstrequal{#1}{2}{\textcolor{blue}{R#1}}{%
        \ifstrequal{#1}{3}{\textcolor{magenta}{R#1}}{%
        \ifstrequal{#1}{4}{\textcolor{teal}{R#1}}{%
                           \textcolor{cyan}{R#1}%
        }}}}%
    }%
}}

\usepackage{xr-hyper}

\makeatletter
\newcommand*{\addFileDependency}[1]{
  \typeout{(#1)}
  \@addtofilelist{#1}
  \IfFileExists{#1}{}{\typeout{No file #1.}}
}

\makeatother

\definecolor{cvprblue}{rgb}{0.21,0.49,0.74}
\usepackage[pagebackref,breaklinks,colorlinks,allcolors=cvprblue]{hyperref}
\usepackage[capitalize]{cleveref}
\crefname{section}{Sec.}{Secs.}
\crefname{table}{Table}{Tables}
\crefname{figure}{Fig.}{Figs.}

\ifarxiv \crefname{appendix}{App.}{Apps.}
\else \crefname{appendix}{Suppl.}{Suppls.} \fi

\begin{document}

\title{\paperTitle}
\author{\authorBlock}
\maketitle

\begin{abstract}
Collaborative perception enables vehicles to overcome individual perception limitations by sharing information, allowing them to see further and through occlusions. In real-world scenarios, models on different vehicles are often heterogeneous due to manufacturer variations. Existing methods for heterogeneous collaborative perception address this challenge by fine-tuning adapters or the entire network to bridge the domain gap. However, these methods are impractical in real-world applications, as each new collaborator must undergo joint training with the ego vehicle on a dataset before inference, or the ego vehicle stores models for all potential collaborators in advance. Therefore, we pose a new question: Can we tackle this challenge directly during inference, eliminating the need for joint training? To answer this, we introduce Progressive Heterogeneous Collaborative Perception (PHCP), a novel framework that formulates the problem as few-shot unsupervised domain adaptation. Unlike previous work, PHCP dynamically aligns features by self-training an adapter during inference, eliminating the need for labeled data and joint training. Extensive experiments on the OPV2V dataset demonstrate that PHCP achieves strong performance across diverse heterogeneous scenarios. Notably, PHCP achieves performance comparable to SOTA methods trained on the entire dataset while using only a small amount of unlabeled data. Code is available at: \url{https://github.com/sihaoo1/PHCP}
\end{abstract}
\section{Introduction}
Recent advances in Vehicle-to-Everything (V2X) have improved the capability of autonomous driving systems\cite{zhao2020enhanced,yusuf2024vehicle, kitajima2022nationwide}. V2X enables vehicles to interact with the surrounding environment, creating a more intelligent and interconnected traffic ecosystem. With the support of V2X, Collaborative Perception allows vehicles to share perception information\cite{thunberg2021efficiently,xie2022safe,shladover2021opportunities}, thus expanding the range of perception and empowering the ability to see through occlusion\cite{han2023collaborative}. This achieves a more precise and robust environmental perception in complex urban traffic environments.

Despite the potential of collaborative perception in improving the performance of autonomous driving systems, it faces a major challenge in the real world: heterogeneity\cite{li2024breaking,shao2024hetecooper,yazgan2024survey, xu2023bridging,luo2024plug, lu2024extensible}. Autonomous vehicles produced by different manufacturers typically adopt different sensor configurations and perception models, resulting in significant differences in the encoded intermediate features in the semantic space (domain gap)\cite{xu2023bridging}. This problem makes collaborative perception between intelligent agents difficult and may lead to ineffective information sharing and feature fusion.

\begin{figure}[t]
    \centering
    \begin{subfigure}[b]{0.48\linewidth}
        \centering
        \includegraphics[width=\linewidth]{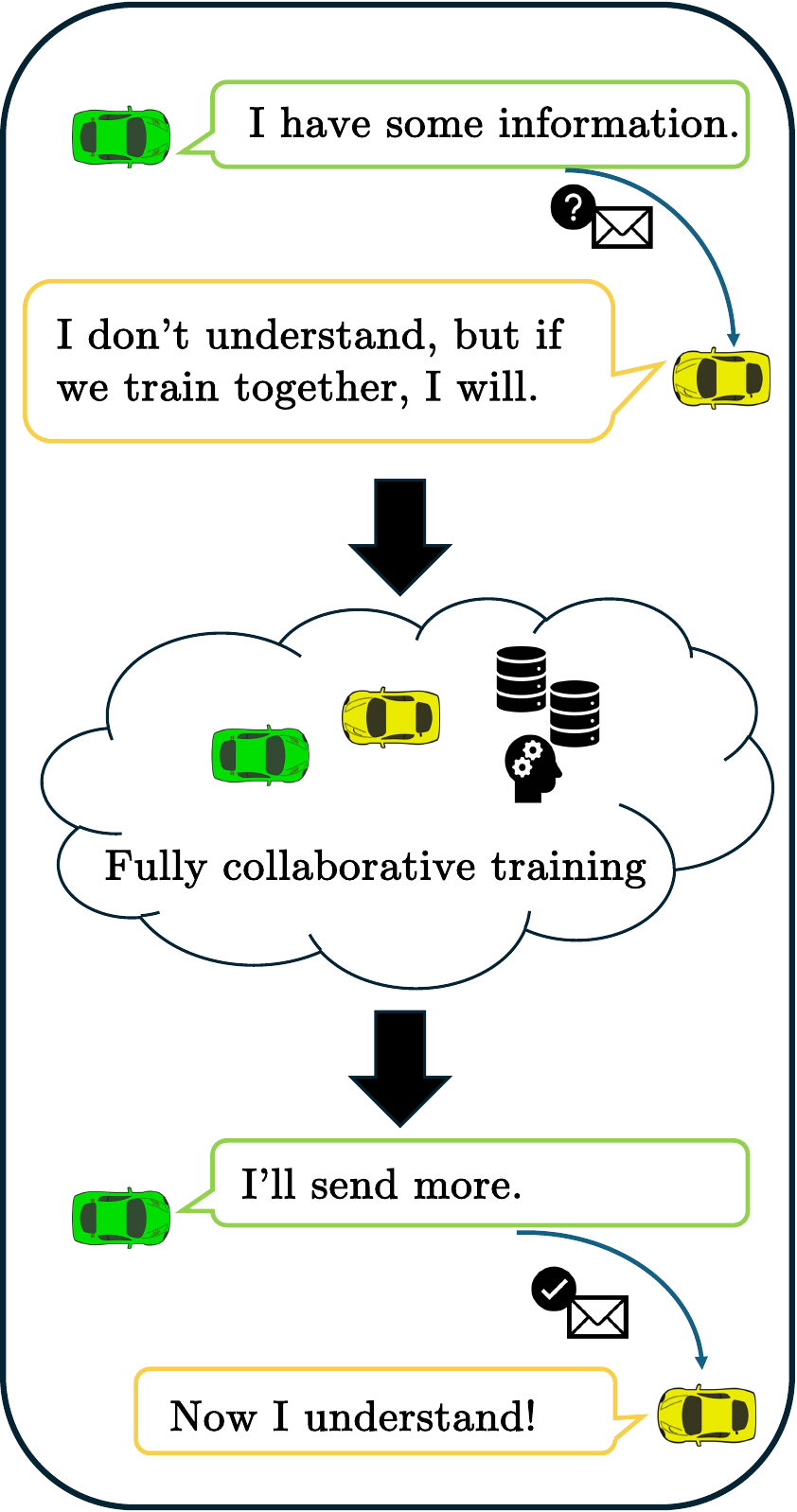}
        \caption{}
        \label{fig:other_method}
    \end{subfigure}
    \hfill
    \begin{subfigure}[b]{0.48\linewidth}
        \centering
        \includegraphics[width=\linewidth]{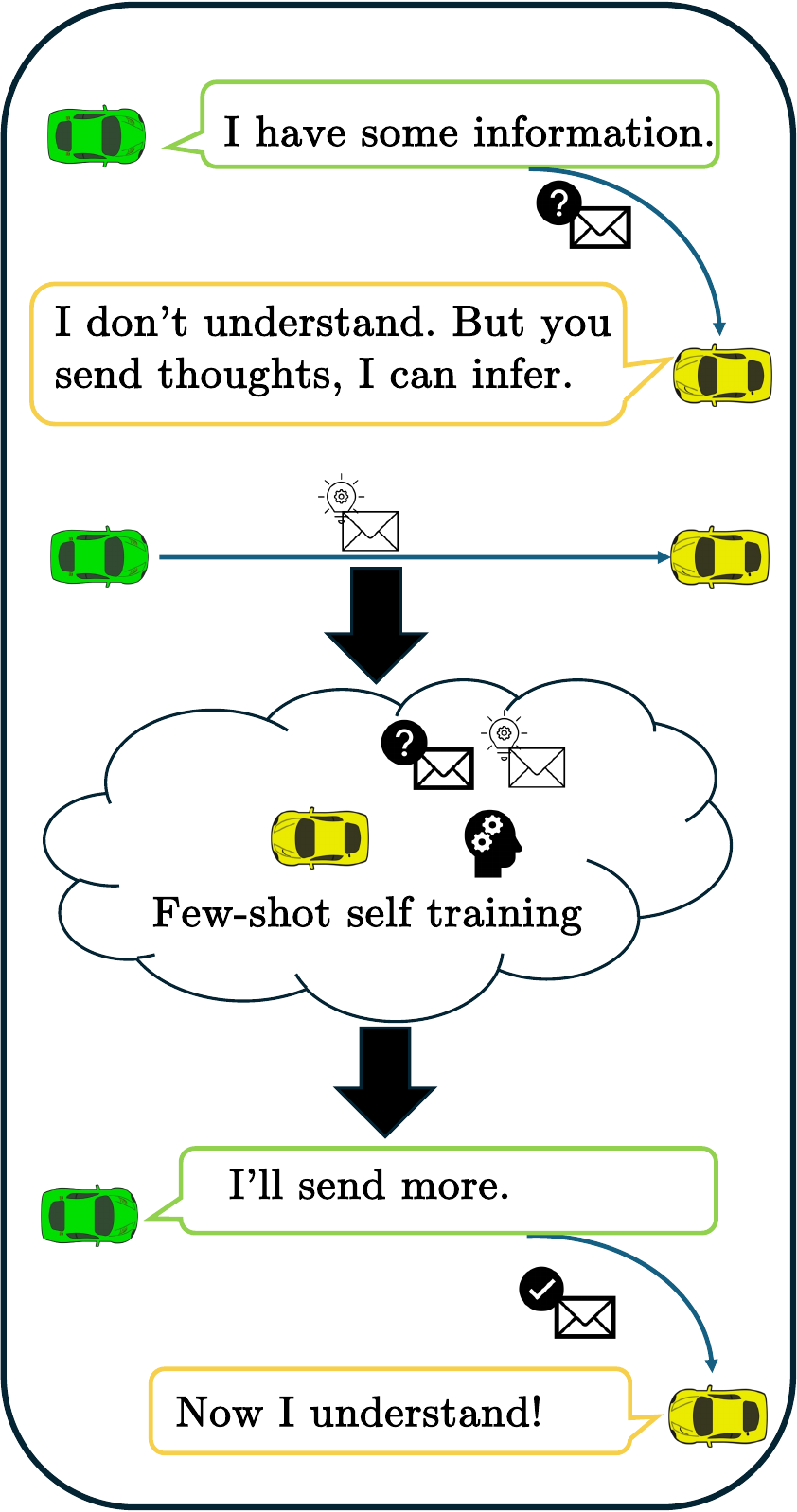}
        \caption{}
        \label{fig:my_method}
    \end{subfigure}
    \caption{Illustration of different heterogeneous collaborative perception patterns. Fig.\ref{fig:other_method} corresponds to existing methods requiring joint training before collaboration. Fig.\ref{fig:my_method} shows our PHCP process. It utilizes a small amount of unlabeled data from other agents to complete self-training and collaborate with others.}
\end{figure}

Many existing methods enhance heterogeneous collaborative perception by training feature encoders and detection heads\cite{lu2024extensible, xu2023bridging} or by learning adapters\cite{luo2024plug} for new collaborators. While these approaches improve performance, they remain limited in their generalization to newly added agents. As shown in Fig.\ref{fig:other_method}, this limitation implies that whenever a new agent joins the collaboration, the model or adapter must be trained, and additional parameters must be stored. However, with the continuous emergence of new sensors and perception models, it becomes impractical for autonomous vehicles to maintain dedicated adapters or models for every potential collaborator. This highlights the need for a more flexible and scalable heterogeneous collaborative perception framework. To address this challenge, we propose a new question: \textbf{Can the ego vehicle dynamically change its model parameters during the inference stage for different collaborators without relying on joint training or any prior knowledge of them?} 

Solving this problem presents several key challenges: (1) Lack of supervision in the inference stage. Unlike the training stage, the inference process lacks additional annotations. Therefore, a core challenge is leveraging unsupervised learning methods to fine-tune the model and achieve feature alignment. (2) Constraints on data volume and computational time. Given the real-time requirements of collaborative perception, we aim to minimize both the training data amount and fine-tune iterations after the collaboration relationship is confirmed. 

To address these challenges, we propose a Progressive Heterogeneous Collaborative Perception(PHCP) framework based on self-training and few-shot unsupervised domain adaptation. This framework aligns features in semantic space by only a small amount of unlabeled data from collaborating agents in the inference stage. The overall workflow consists of two stages. In stage I, the agent generates pseudo labels from its predictions and transmits them with intermediate features to the ego vehicle. Upon receiving this information, the ego vehicle fine-tunes an adapter by a few-shot learning approach. In stage II, the agent transmits only the intermediate features, while the ego vehicle uses the well-trained adapter to transform the features and then perform fusion and prediction. We conduct extensive experiments on the OPV2V dataset, and the results demonstrate that our method outperforms the direct collaboration baseline by approximately 30\% in heterogeneous collaborative perception scenarios. Our main contributions are summarized as follows:
\begin{itemize}[leftmargin=15pt]
    \vspace{8pt}
    \item To the best of our knowledge, this is the first work to tackle the domain gap in heterogeneous collaborative perception during the inference process.
    \vspace{8pt}
    \item We provide a new perspective to address this challenge through few-shot unsupervised domain adaptation. We propose a novel collaborative perception scheme and framework, generating effective supervision labels during inference and enabling the ego vehicle to fine-tune the adapter and resolve the domain gap issues within a few iterations,
    \vspace{8pt}
    \item We conducted extensive experiments across various scenarios, demonstrating the robustness of our method in heterogeneous collaborative perception.
    
\end{itemize}

\label{sec:intro}

\section{Related Work}
\label{sec:related}

\begin{figure}[t]
    \centering
    \includegraphics[width=\linewidth]{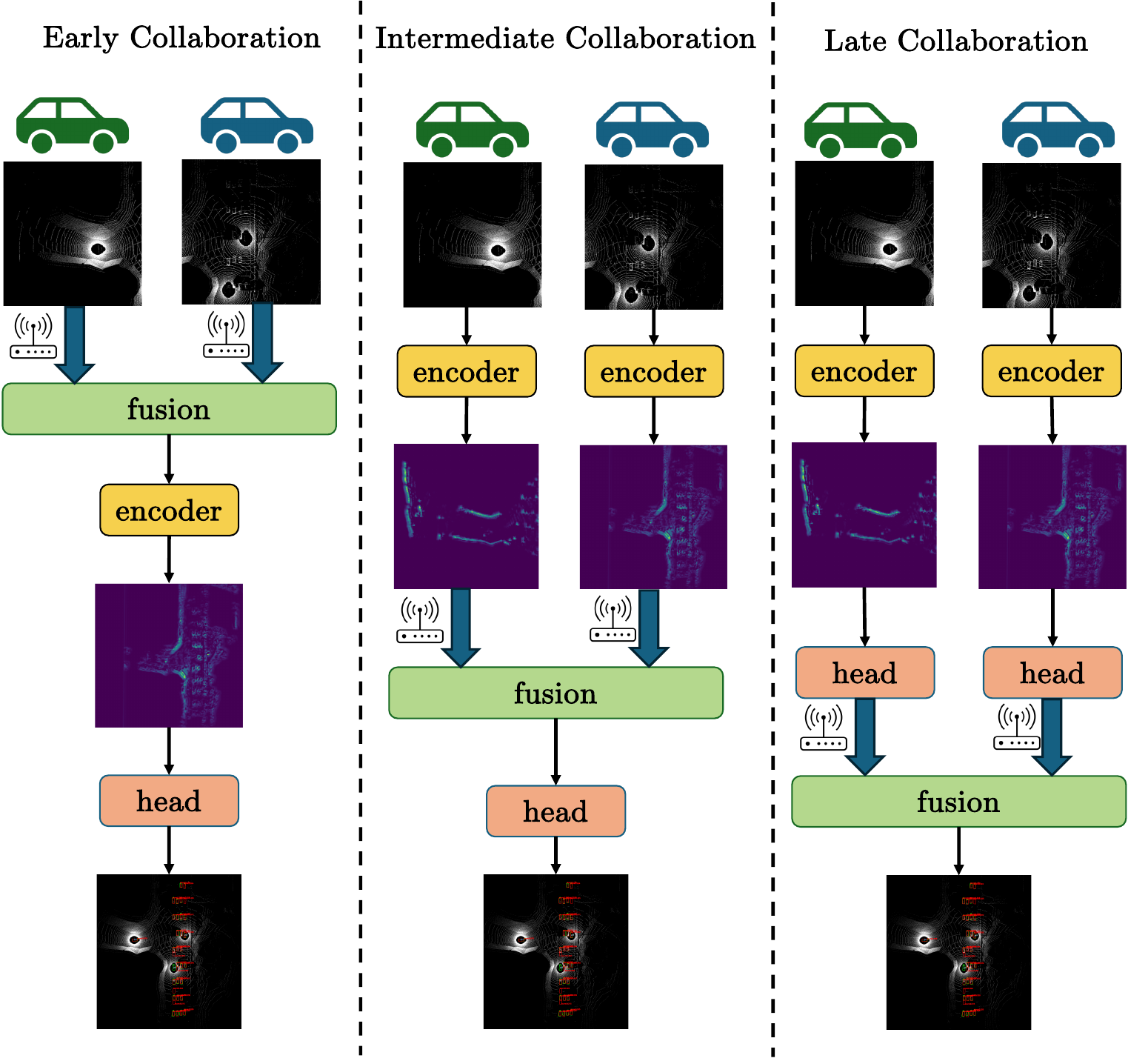}

    \caption{Comparison of different collaborative perception schemes.}
    \label{fig:fusion_comparison}
\end{figure}

\textbf{Collaborative perception.} Collaborative perception improves perception performance by fusing shared information among connected vehicles. As shown in the Fig.\ref{fig:fusion_comparison}, collaborative perception can be categorized into early, late, and intermediate collaboration depending on the type of shared data\cite{han2023collaborative}. In early collaboration, raw sensor data such as LiDAR point clouds and RGB images are directly fused at the data level after calibration and spatial alignment, resulting in a more comprehensive environmental perception\cite{gao2018early_object,chen2019early_cooper, arnold2020early_cooperative}. Although this method performs well, the transmission of large amounts of raw data from the vehicle poses challenges to the V2X network. In late collaboration, object-level fusion is used, where agents only share perception results. After post-processing steps like spatial transformation and non-maximum suppression, the final detection results are generated\cite{fu2020late_depth, zeng2020late_dsdnet, chen2022late_learning, su2023late_uncertainty}. This method is simple to implement and has low bandwidth requirements but is more sensitive to noise and localization errors, leading to relatively lower performance. Intermediate collaboration is a feature-level fusion, gaining increasing attention in recent years. It transmits feature-level data while sharing and preserving semantic information, effectively reducing data transmission volume\cite{liu2020inter_when2com, hu2022inter_where2comm, wang2023umc}. It’s a trade-off solution between early and late collaboration. Recent studies\cite{li2023inter_multi,lu2023inter_robust,wang2023inter_core,yang2023inter_spatio, cui2022inter_coopernaut, li2021disconet, hu2023coca3d} have focused on this scheme. F-Cooper\cite{chen2019fcooper} is the pioneering work in this field, introducing a low-level voxel fusion approach and a spatial feature fusion strategy. V2VNet\cite{wang2020v2vnet} leverages graph neural networks to model agent communication, achieving a balance between accuracy and bandwidth requirements.  AttFusion\cite{attfusion} introduces a single-head self-attention fusion module to capture spatial relationships within feature maps. CoBEVT\cite{xu2022cobevt} explores an alternative approach by replacing LiDAR with cameras, proposing a fused axial attention module that effectively integrates camera-based BEV features, enhancing perception performance in a LiDAR-free scenario. However, these methods assume that all agents share the same model and parameters\cite{xu2023bridging,luo2024plug,lu2024extensible}, which is unrealistic in the real world. So, recent studies have shifted their focus toward heterogeneous collaborative perception, aiming to address this challenge.

\noindent \textbf{Heterogeneous Collaborative perception.}V2X-ViT\cite{xu2022v2x} was the first to address the heterogeneity problem. It treats vehicle-to-infrastructure and vehicle-to-vehicle as two fusion types and proposes the Heterogeneous Multi-Agent Attention module to perform fusion among heterogeneous agents. HM-ViT\cite{xiang2023hm} extends heterogeneous collaborative perception to multi-sensor settings, introducing a heterogeneous 3D graph attention module that effectively fuses BEV features from different modalities. MPDA\cite{xu2023bridging} focuses on model-level heterogeneity and employs a learnable feature resizer to align features, along with a sparse cross-domain transformer to bridge the domain gap.  PnPDA\cite{luo2024plug} introduces a novel plug-and-play domain adapter, effectively bridging the domain gap without destructing the models. HEAL\cite{lu2024extensible} proposes a new framework that establishes a unified feature space for all agents, ensuring that when a new agent joins the collaboration, it only needs to be aligned to this shared space rather than adapting to every collaborator. Although these methods address the domain gap caused by heterogeneity, they rely on prior training on the dataset and lack the flexibility to adapt to newly introduced agents.

\begin{figure*}[!t]
\centering
    \centering{\includegraphics[width=1\linewidth]{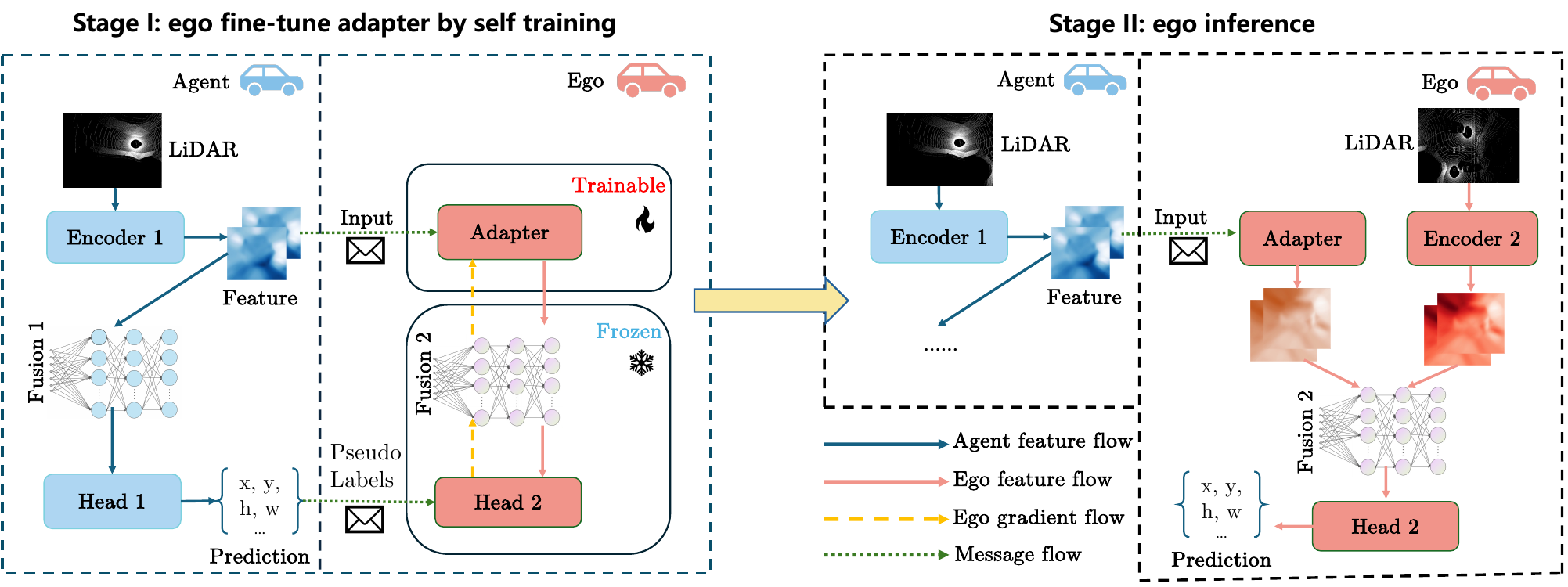}}
\caption{Overview of our proposed heterogeneous collaboration process. In Stage I, the ego performs self-training using pseudo labels to fine-tune the adapter. In Stage II, the ego applies the adapter to transform features and then perform fusion and prediction.}
\label{fig:overview}
\end{figure*}

\section{Method}
\label{sec:method}
In this paper, we explore achieving progressive collaboration in heterogeneous scenarios without relying on joint training or prior knowledge of other agents. Our method first generates pseudo labels from agents to facilitate self-training for the ego. This process is completed within a very limited number of frames. Subsequently, the agent only needs to share intermediate features, while the ego performs feature fusion and final prediction, thereby enhancing collaboration efficiency and addressing the domain gap problem. Fig.\ref{fig:overview} illustrates the main workflow.

\subsection{Problem Formulation}

The collaborative perception at the feature level can be divided into two stages. In the first stage, all agents use their own encoders to process the raw data and generate feature maps, which are then shared. In the second stage, ego vehicles perform fusion on all agents' features, and then the detection head is used to predict the results. Consider N agents in the scene, where $\chi_i$ represents the ith agent. The process can be described as follows:
\begin{align}
    \mathbf{F_i} &= \text{enc}_i (d_i) \label{eq:1},\quad \mathbf{F_i} \in \mathcal{S}\\
    \mathbf{P_i} &= \text{head}_i \left( \text{fuse}_i (\mathbf{F_1}, \mathbf{F_2}, \dots, \mathbf{F_n}) \right) \label{eq:2}
\end{align}
where $d_i$ denotes $\chi_i$'s original data, $P_i$ is the final prediction result. $enc_i$ is feature encoder of $\chi_i$, which outputs the feature map $F_i$. The feature fusion module $fuse_i$ is responsible for spatial-temporal alignment and feature integration. $head_i$ is the classification head for $\chi_i$, predicting the target position and confidence score based on the feature map.
If heterogeneity is considered, the intermediate features $F_i$ generated by the encoding may belong to different semantic spaces. To address this, we can use feature adapters to map the features of different agents to the ego's feature space.
\begin{equation}
    \mathbf{F_i^{\prime}}=\Phi_{i \to ego} (\mathbf{F_i}), \quad \mathbf{F_i^{\prime}} \in \mathcal{S}_{ego}, \mathbf{F_i} \in \mathcal{S}_i 
\end{equation}
Our target is to address the heterogeneity problem by achieving adaptive adjustment of the adapter $\Phi_{i \to ego}() $ during the collaborative perception inference stage with only a small amount of unlabeled data.

\subsection{Feature Adapter}
\begin{figure}[htbp]
    \centering
    \begin{subfigure}[b]{0.45\linewidth}
        \centering
        \includegraphics[width=\linewidth]{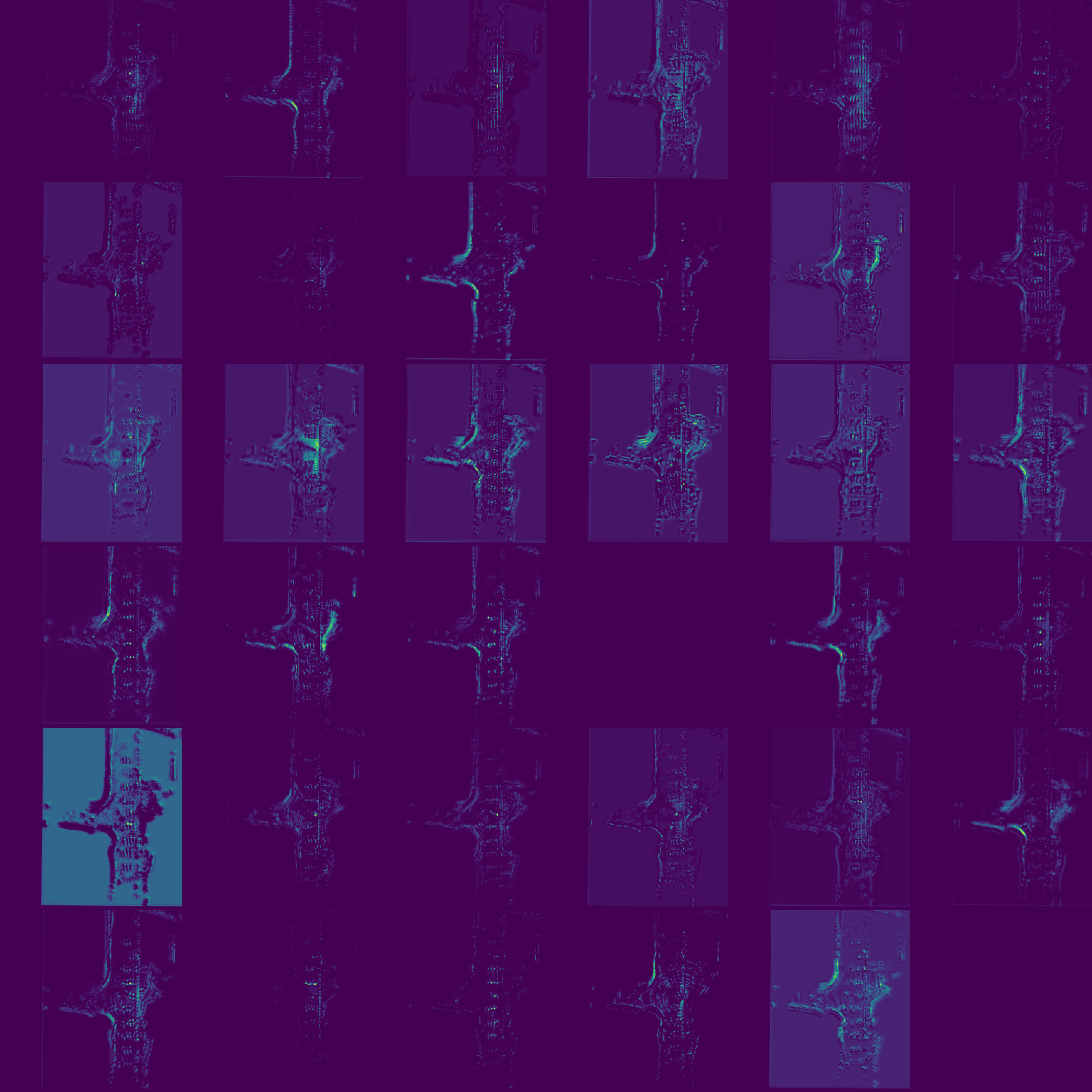}
        \caption{PointPillar feature map}
        \label{fig:image1}
    \end{subfigure}
    \hfill
    \begin{subfigure}[b]{0.45\linewidth}
        \centering
        \includegraphics[width=\linewidth]{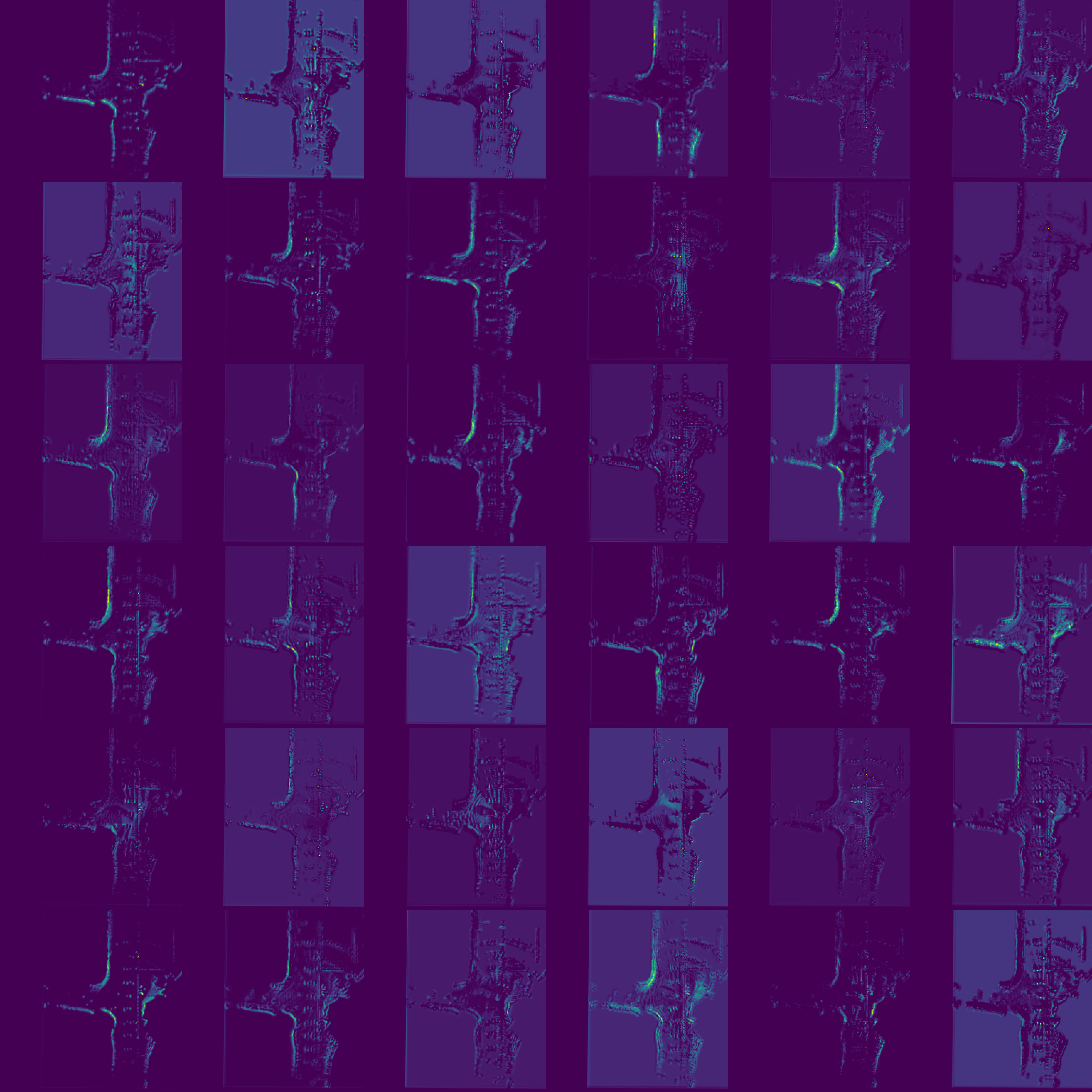}
        \caption{SECOND feature map}
        \label{fig:image2}
    \end{subfigure}
    
    \vspace{0.5em} 
    \begin{subfigure}[b]{0.45\linewidth}
        \centering
        \includegraphics[width=\linewidth]{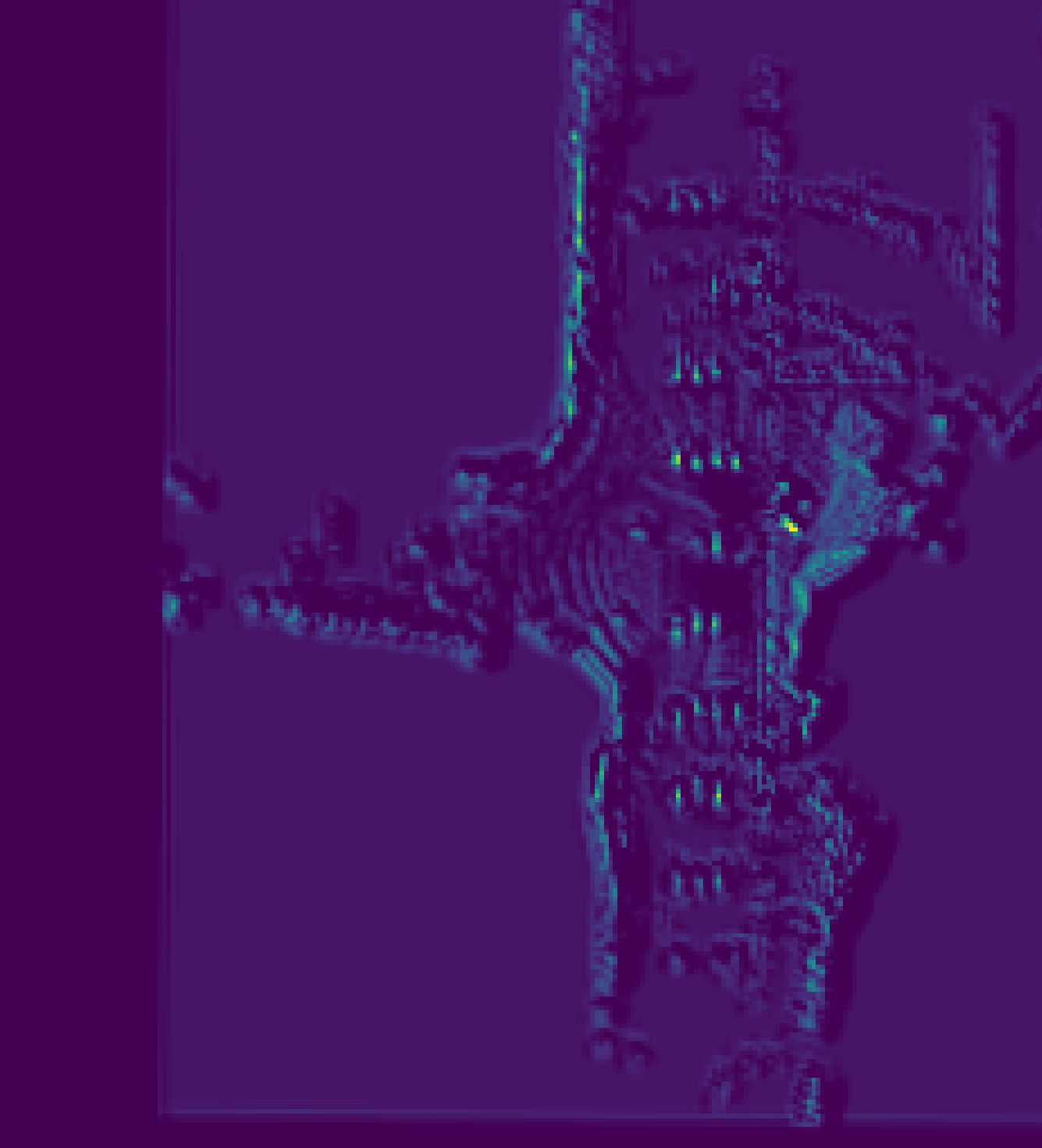}
        \caption{PointPillar feature map, c=29}
        \label{fig:image3}
    \end{subfigure}
    \hfill
    \begin{subfigure}[b]{0.45\linewidth}
        \centering
        \includegraphics[width=\linewidth]{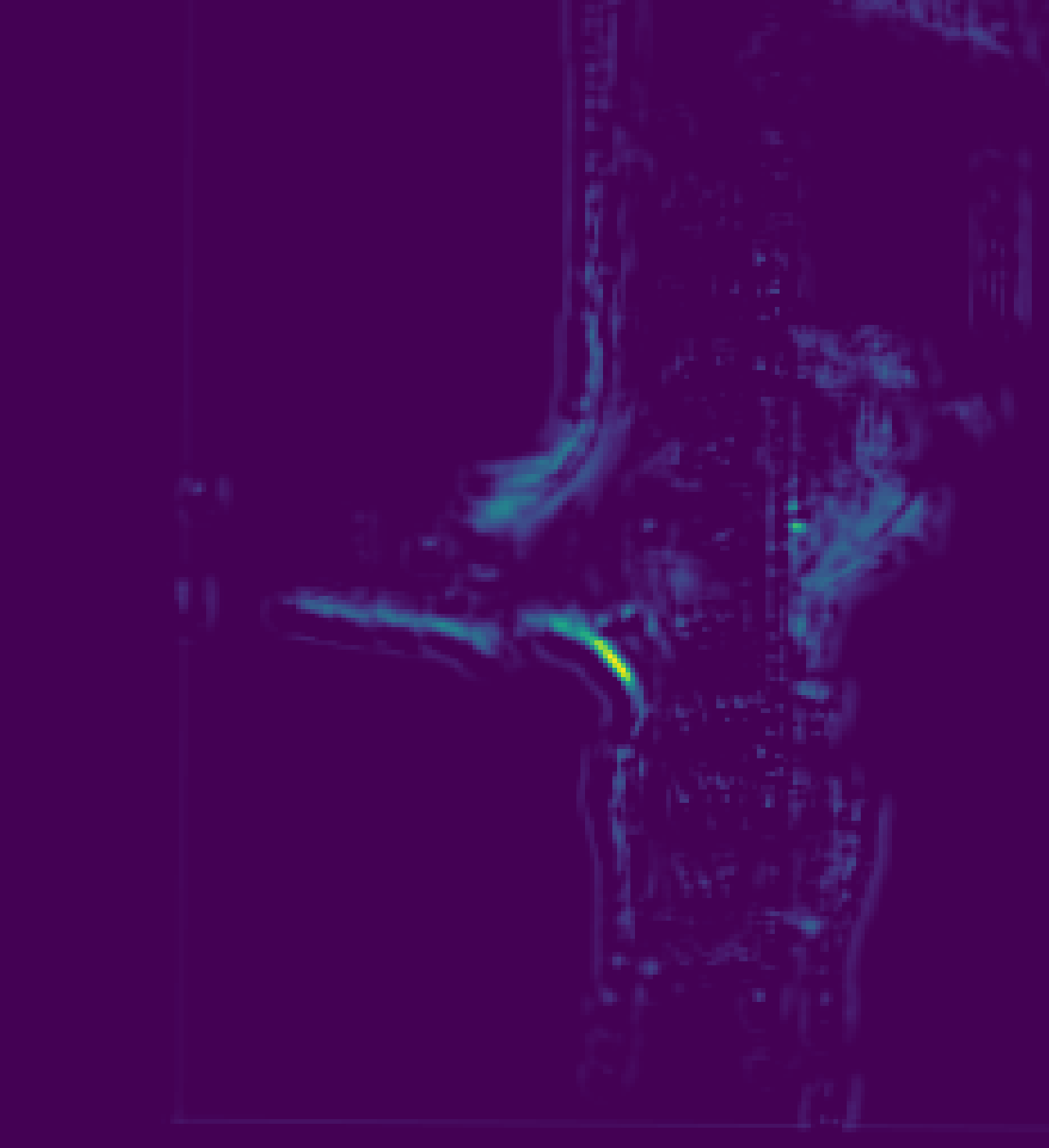}
        \caption{SECOND feature map, c=29}
        \label{fig:image4}
    \end{subfigure}
    \caption{\textbf{Illustration of domain gap of different encoders.}We visualized the encoder's output features by displaying each channel's absolute values. Figure \ref{fig:image1} and Figure \ref{fig:image2} illustrate the misalignment in the channel dimension. Figures \ref{fig:image3} and \ref{fig:image4} explain the differences in the area of interest} 
    \label{fig:feature_map_comp}
    
\end{figure}

Our approach employs a feature adapter to bridge the domain gap between the source $\mathcal{S}_{i}$ and target domains $\mathcal{S}_{ego}$. Since we perform training on a small dataset, the adapter structure must remain simple to prevent overfitting. Through visual analysis of feature maps in Fig.\ref{fig:feature_map_comp}, we observed that while the overall feature distributions of the source and target domains are similar, they show misalignment in channel dimensions and certain critical regions. We found CBAM\cite{woo2018cbam} to be particularly well-suited for this task. CBAM effectively addresses this issue by incorporating both channel and spatial attention mechanisms. The channel attention module(CAM) directs the network to focus on important channels, while the spatial attention module(SAM) enhances attention to crucial spatial regions. CBAM’s lightweight design also introduces minimal computational overhead, making it an efficient and practical choice for our method.

\subsection{PHCP Scheme}
We adopt a novel collaborative perception approach between intermediate and late collaboration, ensuring the generation of pseudo-labeled data without introducing excessive communication overhead. Specifically, after establishing the collaboration relationship, in the first k frames, the agent simultaneously sends features and detection results to the ego,  and the ego uses this information to update the feature adapter. We define this as the stage I. The updated method will be introduced later. After k frames, the collaboration process follows the normal intermediate collaboration process, with the agent only sending features to the ego. This corresponds to stage II. We determine the value of k by referring to typical values in few-shot learning, choosing 1, 5, or 10.

\subsection{PHCP Process}
This progressive collaboration process consists of two stages. In the first stage, the ego trains its own adapter using features and pseudo-labels provided by agents through a few-shot self-training way. In the second stage, the ego aligns the features of the agents by the adapter and then performs fusion and prediction.

\vspace{8pt}
\textbf{Stage I.} We aim to optimize the ego’s adapter without prior knowledge about agents in this stage. Due to the lack of labels, we use agents' predictions as pseudo-labels. In particular, we select high-quality predictions as pseudo-labels based on confidence scores, and we also record the corresponding confidence scores as soft labels instead of one-hot labels. In this way, we can create dataset $\mathcal{D}_i = \{(d_1, p_1), \dots, (d_k, p_k) \mid d \in \mathcal{S}_i \}$ as a small training sample dataset(the support set).
\begin{algorithm}[]
    \caption{\textbf{Stage I: adapter fine-tuning}}\label{alg:fusion}
    \begin{algorithmic}[1]
    
        \For{each scenario}
        \State \textbf{Input: } $\{F_i\},\mathbf{F}_i=\{d_1,..d_k\}$ \Comment{shared feature from each agent}
        \For{each agent $\chi_i$ except ego}
            \State $\mathcal{H}_i=\{\mathbf{F_j} \mid \chi_j \text{ is homogenous to } \chi_i\}$ \Comment{aggregate all agents that are homogeneous to the current}
            \State $\mathbf{P_i}=head_i(fuse_i(\mathcal{H}))$
            \State $\mathcal{D}_i=\{\mathcal{H}_i, \mathbf{P_i} \} $
        \EndFor
        
        \For{each agent $\chi_i$ except ego}
            \State $\mathbf{L_i},\mathcal{H}_i \gets \mathcal{D}_i$
            \State $\mathbf{P_i}=head_{ego}(fuse_{ego}(\mathcal{H}_i))$
            \State $\Phi_{i \to ego} \gets \mathop{\nabla} Loss(\mathbf{L_i}, \mathbf{P_i})$
            
        \EndFor
        \State \textbf{Output: } $\Phi_{i \to ego}$ \Comment{feature adapters from agents to ego} 
        \EndFor
    \end{algorithmic} 
\end{algorithm}

Our model optimization strategy builds upon a two-stage fine-tuning approach (TFA)\cite{wang2020few}due to its simplicity and efficiency. We introduce further enhancements to improve its adaptability within our collaborative perception framework. Considering that the detectors of ego and agent have already been trained separately before the collaboration begins, we only need to focus on the second fine-tuning stage. In the second stage, we fix all the parameters of the fusion network and the detector head and only fine-tune the adapters. This is because the ego needs to use the same fusion network and detector head for different agents. When collaborating with multiple agents simultaneously, fine-tuning the model on multiple datasets may introduce mutual interference, reducing its adaptability to each agent and degrading overall performance. However, the adapter is created independently by the ego for each agent, and by only adjusting the adapter, isolation can be achieved, and interference can be reduced. In the fine-tuning process, we adopt a scheduler with a warm-up mechanism. We select one scenario for hyperparameter tuning, while the remaining scenarios are for testing. We train a total of 20 rounds, and due to the small training data size, the overall computational overhead is very low.

\begin{figure}[t]
    \centering
    \begin{subfigure}[b]{0.47\linewidth}
        \centering
        \includegraphics[width=\linewidth]{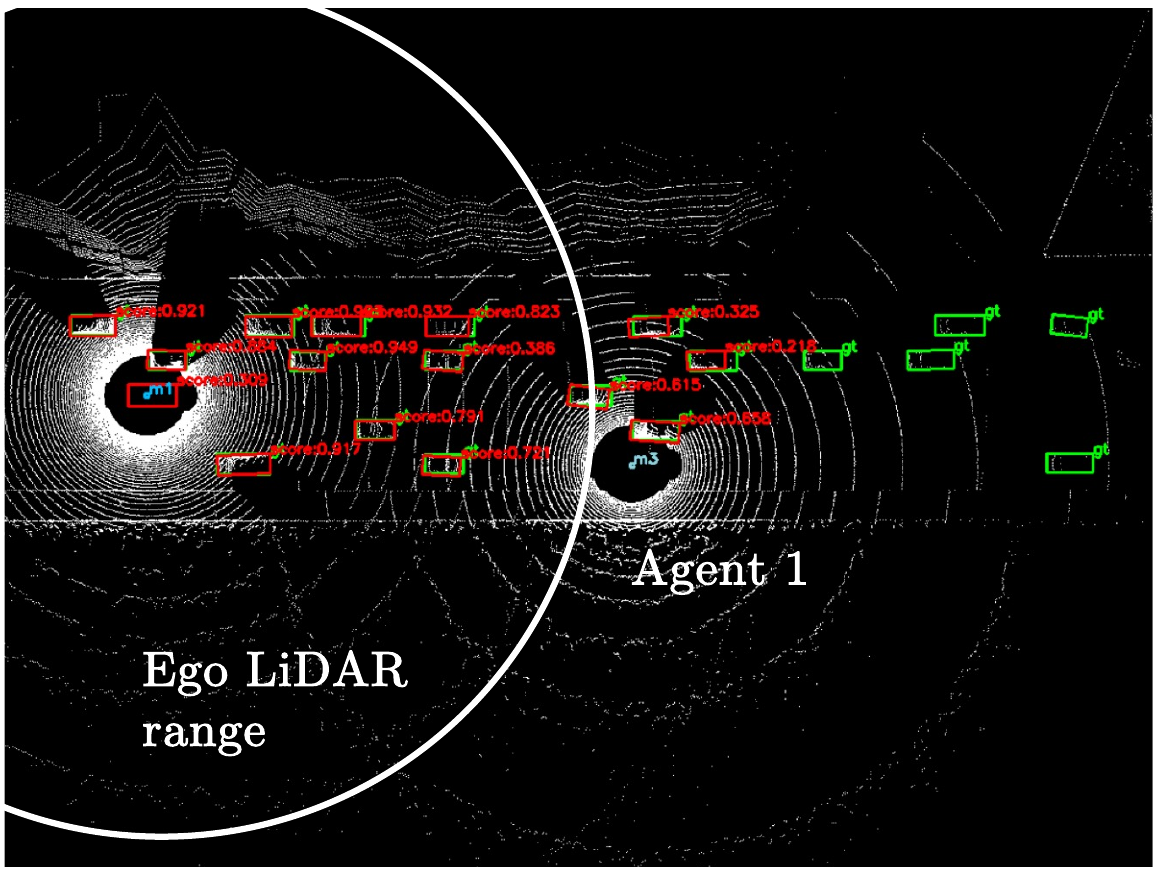}
        \caption{broad shared field of view}
        \label{fig:broad_view}
    \end{subfigure}
    \hfill
    \begin{subfigure}[b]{0.47\linewidth}
        \centering
        \includegraphics[width=\linewidth]{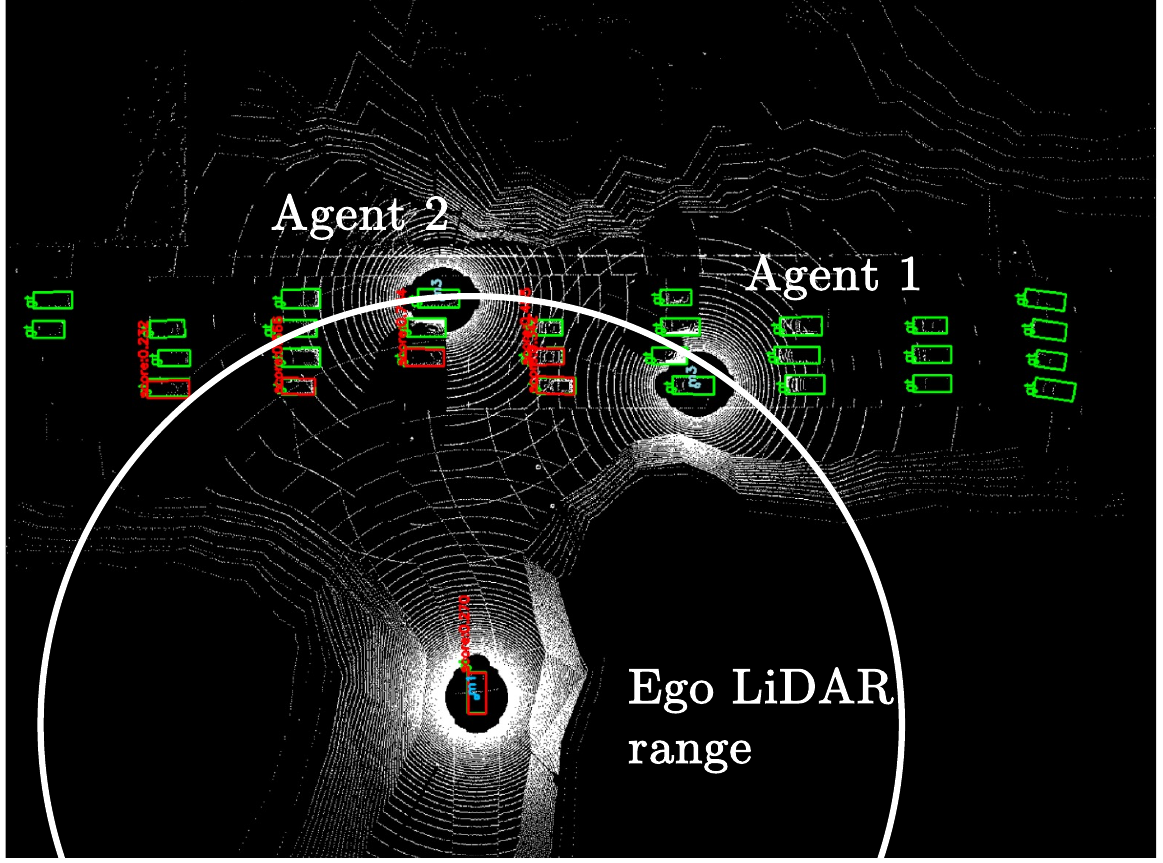}
        \caption{narrow shared field of view}
        \label{fig:narrow_view}
    \end{subfigure}
    \caption{Detection performance under different ego vehicle field of view conditions}
\end{figure}

\vspace{8pt}
\textbf{Stage II.} This stage follows immediately after the first stage.  We consider the features shared by agents as the query set $\mathcal{Q}_i = \{d_{k+1}, \dots, d_N \mid d \in \mathcal{S}_i \}$. We believe that after the training in the first stage, the adapter can effectively address the domain gap problem. Subsequently, the transformed features $\mathcal{Q}_i^{\prime}$ and the features from the ego's encoder can be fused to obtain the final detection results.
\begin{algorithm}[]
    \caption{\textbf{Stage II: Inference}}\label{alg:infer}
    \begin{algorithmic}[1]
    
        \For{each scenario}
        \State \textbf{Input: } $\{F_i\},\mathbf{F}_i={d_{k+1},...d_N}$ \Comment{shared feature from each agent}
        \For{each agent $\chi_i$ except ego}
            \State $F_i^{\prime}=\Phi_{i \to ego}(F_i)$
            \State $\mathcal{F}_i=\mathcal{F}_i \cup F_{i}^{\prime} $
        \EndFor
        \State $\mathbf{P_{ego}}=head_{ego}(fuse_{ego}(\mathcal{F}_i))$
        \State \textbf{Output: } $\mathbf{P_{ego}}$ \Comment{ego final prediction} 
        \EndFor
    \end{algorithmic} 
    
\end{algorithm}
We use the average AP value of all scenarios as the final evaluation metric, referred to as mean scenario AP (mSAP). This is because we observe significant differences in the data distribution among different scenarios. In Fig.\ref{fig:broad_view}, most of the objects are concentrated on one side of the ego vehicle, and even without receiving information from other agents, the final detection results are not affected. However, in Fig.\ref{fig:narrow_view}, the objects are mainly distributed on the side of agent vehicles. If the ego vehicle cannot effectively utilize the feature information from agents, it may result in many false negatives. Therefore, mSAP, as a comprehensive performance evaluation metric, can more fairly measure the detection capability of the model in different scenarios.
\section{Experiments}
\begin{table*}[t]  
    \renewcommand{\arraystretch}{1.2}
    \centering
    \resizebox{\textwidth}{!}{ 
    \begin{tabular}{l|ccccccccccccccccccccc} 
        \toprule
        Method & \multicolumn{16}{c|}{AP@IoU 0.5 in Scenarios} & \multicolumn{3}{c}{mSAP@IoU} \\ 
        \midrule
        & 1 & 2 & 3 & 4 & 5 & 
          6 & 7 & 8 & 9 & 10 & 
          11 & 12 & 13 & 14 & 15 & 
          16 & \multicolumn{1}{|c}{0.3} & 0.5 & 0.7 \textuparrow\\  
        \midrule

        Direct Fusion & 83.2 & 31.9 & 68.2 & 69.0 & 83.9 & 54.7 & 54.3 & 86.8 & 59.9 & 35.0 & 40.2 & 40.5 & 34.0 & 55.4 & 82.0 & 72.5 & \multicolumn{1}{|c}{59.7} & 59.5 & 53.0 \\  

        \textbf{PHCP(Ours)} & \textbf{96.2} & \textbf{98.4} & \textbf{96.2} & \textbf{94.2} & \textbf{96.3} & \textbf{95.2} & \textbf{82.6} & \textbf{97.0} & \textbf{95.2} & \textbf{95.9} & \textbf{88.9} & \textbf{96.2} & \textbf{80.6} & \textbf{85.6} & \textbf{90.9} & \textbf{89.0} & \multicolumn{1}{|c}{\textbf{92.9}} & \textbf{92.4} & \textbf{85.9} \\
        
        \bottomrule
    \end{tabular}
    }
    
    \caption{Comparison between our methods and the direct fusion baseline}
    \label{tab:vs_baseline}
\end{table*}

\begin{table*}[t]  
    \renewcommand{\arraystretch}{1.2}
    \centering
    \resizebox{\textwidth}{!}{ 
    \begin{tabular}{l|ccccccccccccccccccccc} 
        \toprule
        Method & \multicolumn{16}{c|}{AP@IoU 0.7 in Scenarios} & \multicolumn{3}{c}{mSAP@IoU} \\ 
        \midrule
        & 1 & 2 & 3 & 4 & 5 & 
          6 & 7 & 8 & 9 & 10 & 
          11 & 12 & 13 & 14 & 15 & 
          16 & \multicolumn{1}{|c}{0.3} & 0.5 & 0.7 \textuparrow\\  
        \midrule
         F-Cooper\cite{chen2019fcooper} & 84.1 & 50.8 & 69.7 & 56.3 & 79.4 & 65.5 & 57.0 & 76.1 & 61.2 &57.6 & 58.0 & 60.9 & 48.4 & 49.6 & 72.6 & 77.3  & \multicolumn{1}{|c}{91.2} & 83.1 & 63.4 \\
         
        CoBEVT\cite{xu2022cobevt} & 78.7 & 67.2 & 79.5 & 71.8 & 82.0 & 71.2 & 62.2 & 87.4 & 70.3 & 71.2 & 66.4 & 60.4 & 54.1 & 66.8 & 83.9 & 79.7 & \multicolumn{1}{|c}{94.8} & 91.2 & 72.0 \\
        
        AttFusion\cite{attfusion} & 84.0 & 68.5 & 87.0 & 71.4 & 92.6 & 78.6 & 63.7 & 84.0 & 72.8 & 73.5 & 74.4 & 76.3 & 65.1 & 75.7 & 83.7 & 84.9& \multicolumn{1}{|c}{93.0} & 89.9 & 77.3 \\
        
        V2X-ViT\cite{xu2022v2x} & 81.7 & 74.9 & 90.8 & 79.2 & 88.7 & 84.8 & 65.7 & 92.1 & 90.1 & 81.3 & 78.4 & 87.6 & 73.1 & 83.4 & 90.6 & 82.6 & \multicolumn{1}{|c}{94.3} & 92.3 & 82.8\\  
        
        \textbf{PHCP(Ours)} & 89.2 & 92.6 & 90.4 & 87.4 & 92.2 & 94.1 & 67.1 & 95.2 & 89.6 & 94.7 & 84.7 & 91.5 & 70.7 & 83.8 & 89.0 & 80.8& \multicolumn{1}{|c}{92.8} & 92.3 & 87.1 \\

         HEAL\cite{lu2024extensible} & 89.1 & 93.4 & 97.6 & 91.3 & 93.6 & 96.6 & 76.2 & 97.2 & 93.0 & 97.5 & 89.7 & 94.5 & 86.9 & 87.7 & 93.1 & 90.0 & \multicolumn{1}{|c}{95.5} & 95.1 & 91.7 \\
        \bottomrule
    \end{tabular}
    }
    
    \caption{Comparison between PHCP and other methods. Notably, we achieve these results using only a small amount of unlabeled data. }
    \label{tab:vs_others}
\end{table*}

\subsection{Datasets and Evaluation}
\label{sec:datasets}
We use OPV2V\cite{attfusion} and its supplementary dataset, OPV2V-H\cite{lu2024extensible} as the dataset for model training and evaluation. OPV2V is collected by the simulation tools OpenCDA\cite{xu2021opencda} and CARLA\cite{dosovitskiy2017carla}, covering various complex driving scenarios, including challenging situations such as severe occlusion. We use the default dataset splitting ratio. The training and validation sets are used to train the baseline model for homogeneous collaborative perception, and the test set is used to evaluate the performance of heterogeneous collaborative perception.  We further divide the test dataset according to the scenario settings, resulting in 16 specific scenarios. For each scenario, we split them into support sets $\mathcal{S}$ and query sets $\mathcal{Q}$, and the sizes are determined by the number of shots. We adopt mSAP at Intersection-over-Union (IoU) thresholds of 0.3, 0.5, and 0.7  to evaluate the model performance.

\subsection{Experimental Setups}
\label{sec:experiment_setup}
We adopt two types of agents with different LiDAR encoders: agent LP using the PointPillars\cite{lang2019pointpillars} encoder and agent LS using the SECOND\cite{yan2018second} module. Feature fusion is performed using the multi-scale pyramid fusion method proposed by HEAL\cite{lu2024extensible}. The overall training and validation procedure of the model is as follows: on the OPV2V dataset, we conducted homogeneous collaborative perception training for LP and LS respectively using two RTX 6000ADA GPUs, so we get base models $\mathcal{M}_{LP}=\{Enc_{LP}, Fuse_{LP}, Head_{LP}\}$ and $\mathcal{M}_{LS}=\{Enc_{LS}, Fuse_{LS}, Head_{LS}\}$. Subsequently, on the dataset $\mathcal{S}$, we extract intermediate features by $Enc_{LS}$ and generate prediction results using model $Head_{LS}$ as pseudo-labels. Then, we perform fine-tuning on the sub-model $\{\Phi()_{LS \to LP},Fuse_{LP},Head_{LP}\}$ with this pseudo-labled data. Only the adapter is trainable, while the parameters of other modules are fixed. Finally, we combine the adapter with $M_{LP}$ to form a new model $\mathcal{M}_{LP}^{\prime}=\{Enc_{LP}, \Phi()_{LS \to LP},Fuse_{LP}, Head_{LP}\}$, and then evaluated it on the dataset $\mathcal{Q}$.

\noindent \textbf{Optimization Strategy.} We employ a scheduler that combines the warmup and multi-step decay strategies. In the warmup stage, we use a linear strategy, where the warmup factor was set to 0.001, and the warmup lasted for 8 epochs. During the actual training process, the learning rate decayed at the 12th and 16th epochs with a decay factor of 0.1. Our base learning rate was set to 0.005, the batch size was set to 1, and the training was completed within 20 epochs.

\noindent \textbf{Comparison.} Our direct fusion baseline method uses models LS and LP, which have not undergone joint training, for direct collaborative perception. This setting aligns with real-world heterogeneous collaboration, as we have no prior knowledge of each other's model structures or parameters before collaboration begins. Additionally, we compare our approach with other collaborative perception methods, which were fine-tuned on the training dataset before testing. For models that do not support heterogeneous collaborative perception, such as F-Cooper\cite{chen2019fcooper}, V2X-ViT\cite{xu2022v2x}, AttFusion\cite{attfusion}, and CoBEVT\cite{xu2022cobevt}, we first train the models in homogeneous scenarios and then perform 40 fine-tuning epochs in heterogeneous scenarios. For models that support heterogeneous collaborative perception, like HEAL\cite{lu2024extensible}, we follow the training methods provided by the authors.

\subsection{Quantitative evaluation}
\label{sec:quantitative_evaluation}
\textbf{Main performance comparison.} Tab.\ref{tab:vs_baseline} presents a comparison between our method and the direct fusion baseline across various scenarios. Our method consistently outperforms the baseline by 33.2\%, 32.9\%, and 32.9\% in AP@0.3, AP@0.5, and AP@0.7. Our method still achieves strong results even in some challenging scenarios where the baseline performs poorly. Tab.\ref{tab:vs_others} further compares our approach with other collaborative perception methods. Except for HEAL, which achieves the SOTA performance, our method surpasses all others. It is important to note that while other methods utilize the full training dataset with labels, our approach achieves comparable performance using only a minimal amount of unlabeled data.

\noindent \textbf{Computational cost.} As detailed in Tab.\ref{tab:cost}, the training phase takes between 1.5 and 2.4 seconds depending on the number of shots. The inference takes 0.07 seconds. Importantly, the few-shot training process is executed only once when a new collaborative relationship is established. So the overall computational cost meets the real-time requirements.
\begin{table}[H]
\centering
\begin{tabular}{llcc}
\toprule
\textbf{Stage} & \textbf{Setting} & \textbf{Time(s)} & \textbf{Mem(MB)} \\
\midrule
\multirow{6}{*}{Training} 
& shots=1, bs=1 & 1.49 & 1290 \\
& shots=2, bs=2 & 1.71 & 2314 \\
& shots=4, bs=4 & 2.17 & 4552 \\
& shots=5, bs=5 & 2.39 & 5604 \\
\midrule
\multirow{1}{*}{Inference} 
& - & 0.07 & 798 \\
\bottomrule
\end{tabular}
\caption{Computation cost of training and inference. All training configurations use 20 iterations; \textbf{shots} indicates the number of shots used for training, and \textbf{bs} is the batch size.}
\label{tab:cost}
\end{table}

\subsection{Qualitative results}
\label{sec:qualitative_results}
\begin{figure*}[htbp]
     \centering{\includegraphics[width=1\linewidth]{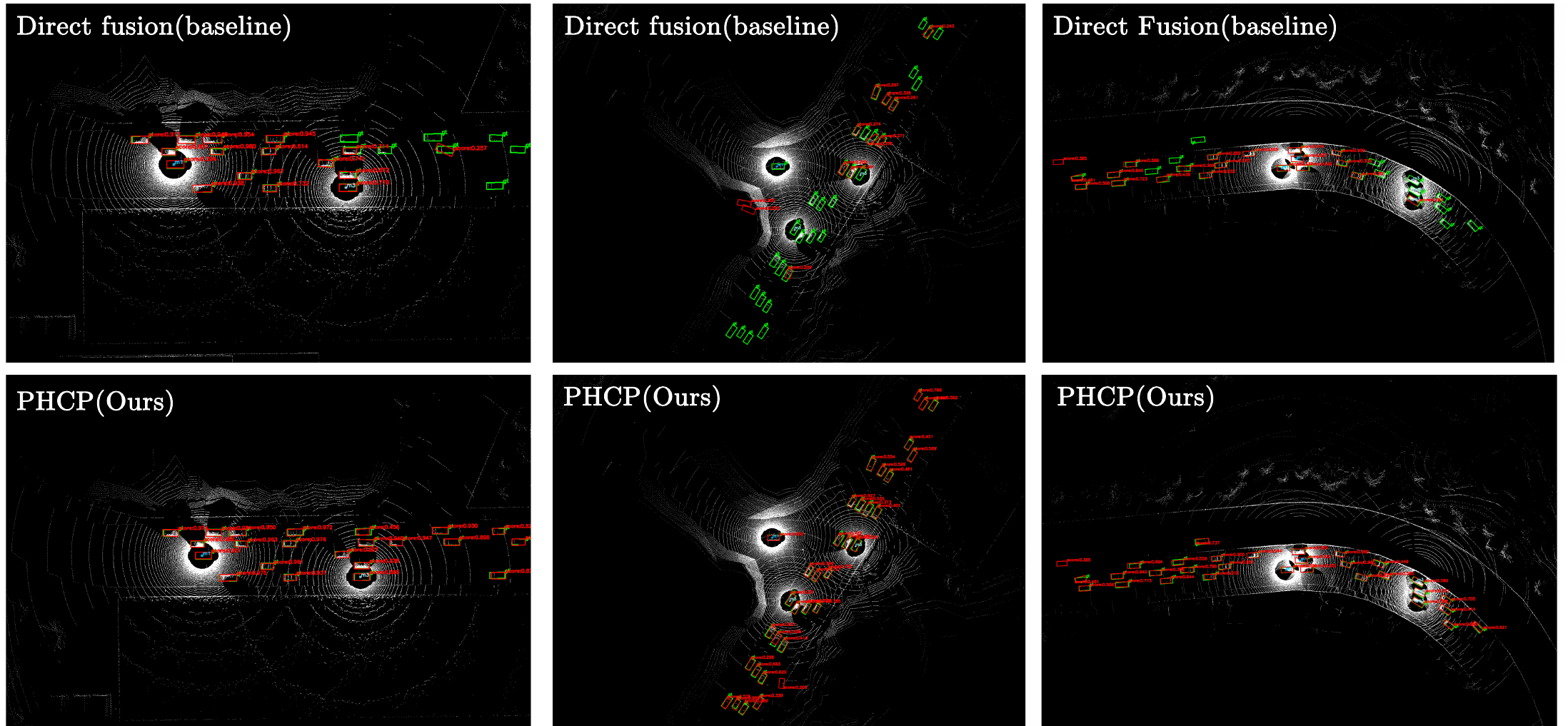}}
     \caption{Visual comparison of three scenarios between our method and the baseline. Each scenario involves two, three, and four collaborative agents participating in collaborative perception. The \textcolor{red}{red} and \textcolor{green}{green} boxes represent the prediction and ground truth.}
     \label{fig:results_vis}
\end{figure*}
\textbf{Visualization of detection results.}
Fig.\ref{fig:results_vis} presents the qualitative visualizations of our method and the baseline. We select three scenarios with varying numbers of collaborative agents. Due to the lack of effective feature fusion, the baseline method has many false negatives outside the ego vehicle's field of view. In contrast, our approach successfully detects these objects.

\begin{figure}[]
    \centering
    \includegraphics[width=\linewidth]{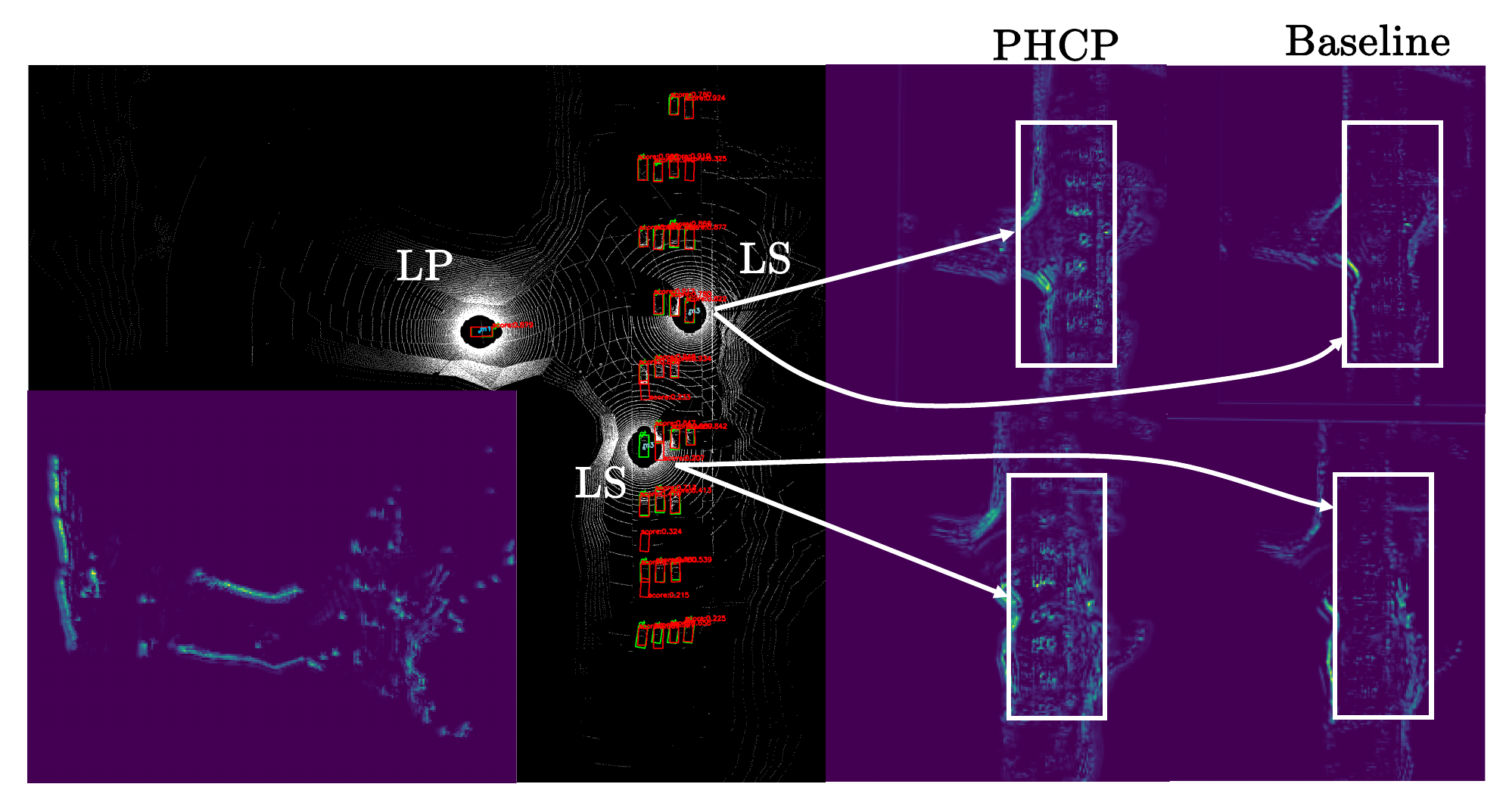}
    \caption{Feature map visualization comparison between our method and baseline}
    \label{fig:fm_out_to_baselinel}
\end{figure}

\noindent \textbf{Visualization of feature map.} Fig.\ref{fig:fm_out_to_baselinel} shows a collaborative scenario involving three agents. By comparing the feature maps, we observe that our method achieves more effective feature alignment with the ego vehicle than the baseline. 

\subsection{Ablation studies}

\begin{figure}[t]
    \centering
    \includegraphics[width=0.88\linewidth]{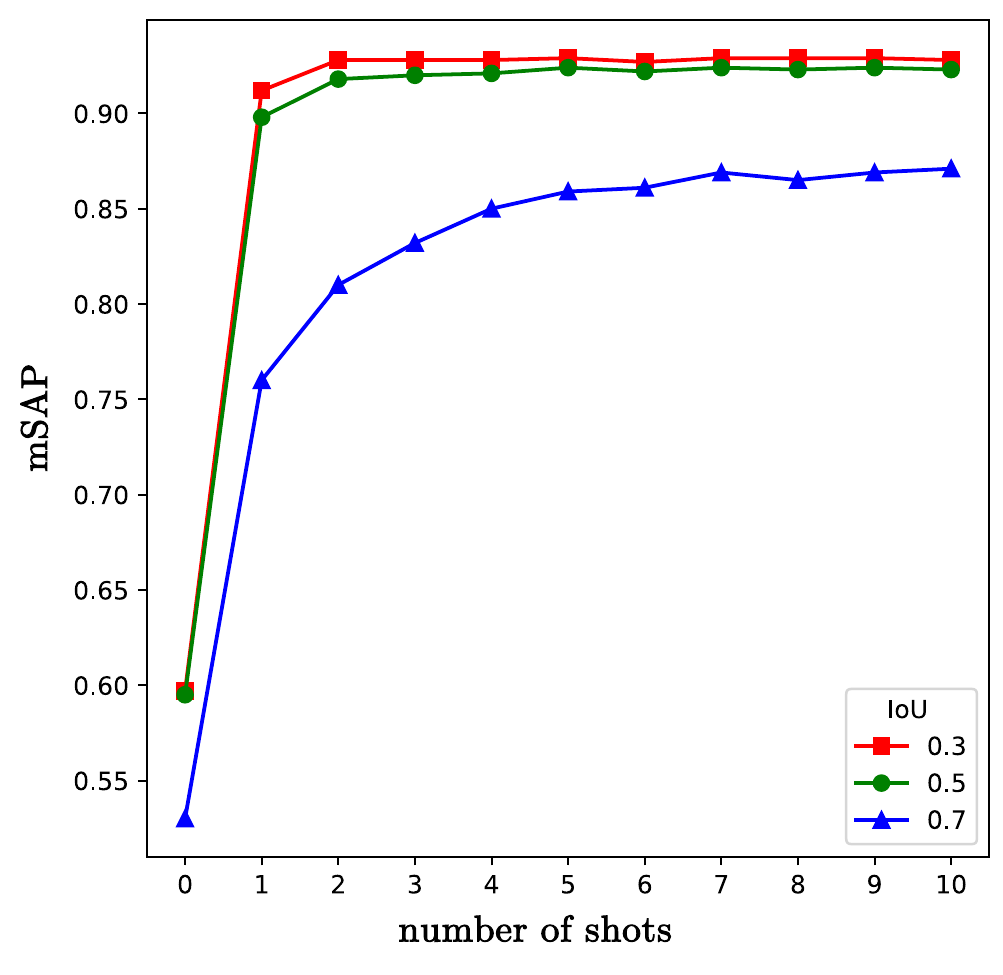}
    \caption{Performance of different number of shots}
    \label{fig:number_of_shots}
\end{figure}

\textbf{Trade-off between performance and number of shots.} A larger number of shots represents more training data, which usually leads to more accurate prediction results. However, it also increases the waiting time for initiating collaboration. Therefore, we analyze the relation between the number of shots and model performance. In our experiments, 0-shot means direct collaboration and serves as the baseline. Fig.\ref{fig:number_of_shots} illustrates the relation between the number of shots and the model's performance. Even when relying only on single-frame data, our model still achieves a performance improvement of approximately 50\%.  As the number of shots increases, the AP@IoU 0.7 also improves. This suggests that more training data can better optimize the adapter module, enhance the quality of fused feature maps, and thus achieve more precise localization and higher prediction accuracy.

\vspace{8pt}
\noindent \textbf{Quality of the pseudo label.} The quality of pseudo labels impacts the final performance. We evaluate the pseudo labels on the test set and obtained an AP@0.5 of 0.86 and an AP@0.7 of 0.78; 80\% of the predictions have confidence scores greater than 0.5. Then we conduct an analysis of different filtering strategies. We set a series of confidence thresholds to filter pseudo-labels or directly used the confidence scores as soft labels. We evaluate the performance using mSAP and AP in the worst-case scenarios(wSAP) for a comprehensive comparison. From Tab.\ref{tab:confidence of label}, retaining only high-quality pseudo labels results in insufficient positive samples, limiting the model’s learning capacity. Conversely, low-quality pseudo labels introduce noise, leading to a decline in overall model accuracy. However, the impact of pseudo-label quality on the final results remains relatively minor. We attribute this to the spatial attention mechanism in the adapter, which effectively focuses on target regions.
\begin{table}[H]
    \centering
    \begin{tabular}{lccccc}
        \toprule
        \ & \ 0.2 & 0.5 & 0.7 & soft \\
        \midrule
        mSAP@0.7 & 85.0 & \textbf{85.9} & 85.8 & 85.4  \\
        wSAP@0.7 & 66.1 & \textbf{68.0} & 67.7 & 67.0 \\
        \bottomrule
    \end{tabular}
    \caption{Performance under different pseudo label confidence levels}
    \label{tab:confidence of label}
\end{table}

\noindent \textbf{Comparison between heterogeneous and homogeneous settings.} From Tab.\ref{tab:msap_iou}, while our baseline model performs well under homogeneous conditions, its performance drops when applied directly to heterogeneous scenarios. This contrast highlights that even strong homogeneous collaboration models cannot effectively handle heterogeneity. In contrast, our method effectively addresses this challenge and maintain robust performance in heterogeneous settings.

\begin{table}[H]
\centering
\begin{tabular}{lccc}
\toprule
\multicolumn{1}{c}{} & \multicolumn{3}{c}{\textbf{mSAP@IoU}} \\
\cmidrule(lr){2-4}
\textbf{Method} & 0.3 & 0.5 & 0.7 \\
\midrule
Direct Fusion(baseline) & 59.7 & 59.5 & 53.0 \\
\textbf{PHCP} & \textbf{92.9} & \textbf{92.4} & \textbf{85.9} \\
SECOND (homogeneous) & 94.8 & 94.2 & 90.5 \\
PointPillar (homogeneous) & 96.2 & 95.8 & 93.1 \\
\bottomrule
\end{tabular}
\caption{Heterogeneous vs. homogeneous collaborative perception.}
\label{tab:msap_iou}
\end{table}

\begin{figure}
    \centering
    \includegraphics[width=\linewidth]{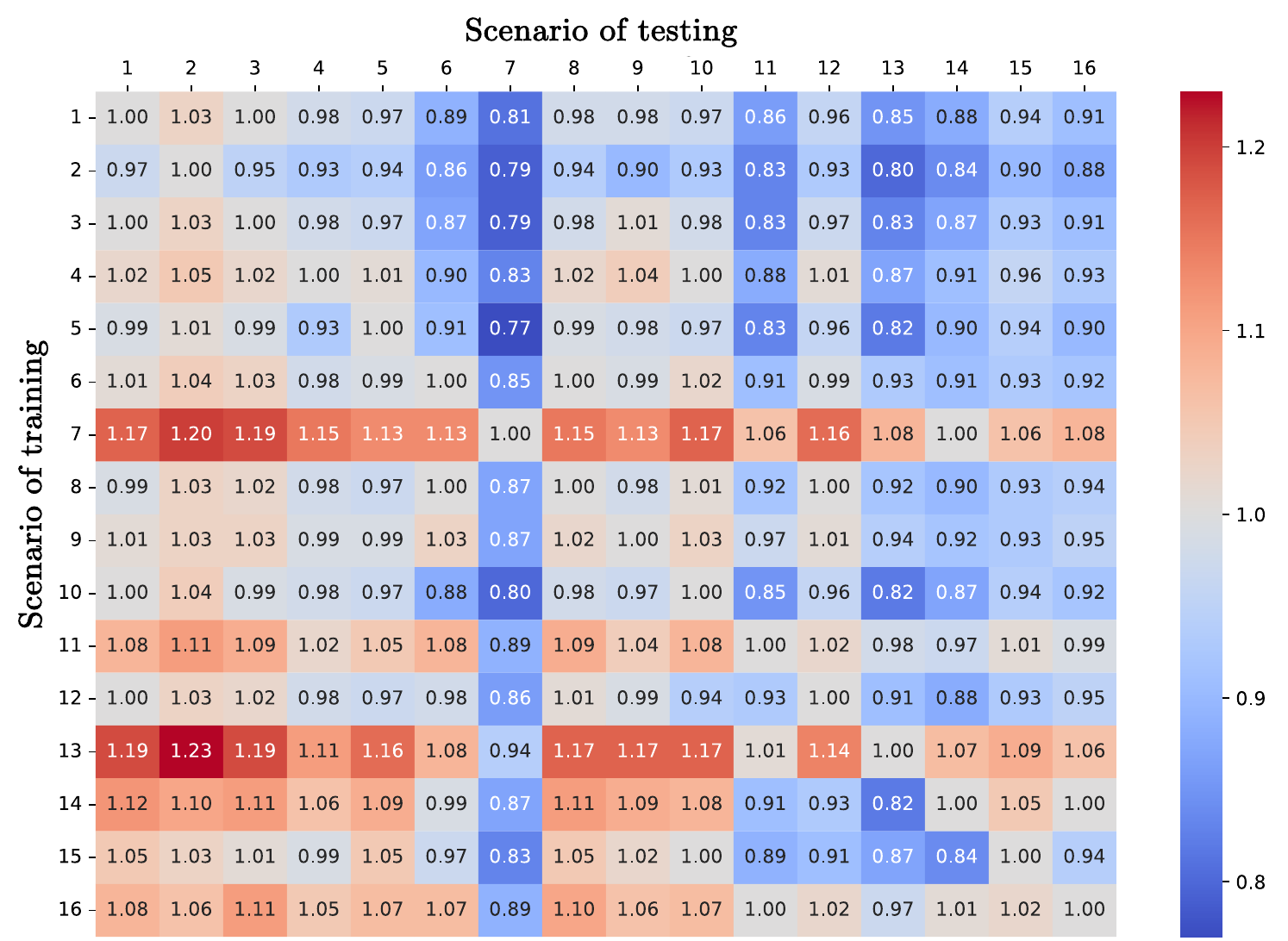}
    \caption{Cross-scenario evaluation results. The figure presents test results across different scenarios. The values have been normalized using the diagonal elements. The overall generalization performance is strong except for a few particularly challenging scenarios where all models perform poorly.}
    \label{fig:heatmap}
\end{figure}

\noindent \textbf{Generalization across scenarios.}
In this study, we conduct separate training and testing on each scenario because each scenario corresponds to a new collaborative relationship, and our model validation aims to assess its adaptability in each new heterogeneous collaboration environment. We also performed cross-scenario testing to verify the model's generalization ability and ensure it does not overfit a single scenario. Specifically, we train a base model for each scenario and used its AP@IoU 0.5 inference results as a baseline. Then, without any additional fine-tuning, we directly test the same model on other scenarios to analyze its generalization ability in different environments. The results in Fig.\ref{fig:heatmap} indicate that although the model's performance slightly declines in some complex scenarios, it remains stable in most scenarios. This result further validates the adaptability and generalization of our model.

\label{sec:Experiments}
\vspace{3pt}
\section{Conclusion}
\label{sec:conclusion}
\vspace{3pt}
We introduce PHCP, a novel heterogeneous collaborative perception framework designed to address feature misalignment during the inference stage of collaborative perception, where training on a dataset with collaborators is unrealistic, and collaborators' prior knowledge is unavailable. The key idea is to utilize pseudo labels generated by other agents and perform few-shot self-training on the ego vehicle to fine-tune the adapter module, thereby bridging the domain gap between the ego and agents. Extensive experiments on the OPV2V dataset demonstrate the effectiveness of our approach, outperforming baseline methods across various heterogeneous collaborative perception scenarios. Compared to other methods trained on the entire dataset, our approach achieves performance comparable to SOTA methods while using only a small amount of unlabeled data.

\newpage
\section*{Acknowledgments}
\label{sec:acknowledgments}
This work was supported by JST, CRONOS, Japan Grant Number JPMJCS24K8.
{\small
\bibliographystyle{ieeenat_fullname}
\bibliography{11_references}
}

\end{document}